\pdfoutput=1
\documentclass[10pt,twocolumn,letterpaper]{article}

\usepackage[usenames,dvipsnames]{xcolor}

\usepackage{iccv}
\usepackage{times}
\usepackage{epsfig}
\usepackage{graphicx}
\usepackage{amsmath}
\usepackage{amssymb}
\usepackage[inline]{enumitem}

\usepackage{xspace}
\usepackage{comment}

\usepackage{changes}
\colorlet{Changes@Color}{orange}

\usepackage{mathtools}
\usepackage{subfig}
\usepackage{makecell}
\usepackage{arydshln}

\DeclarePairedDelimiter\norm{\lVert}{\rVert}%

\newcommand*{\affaddr}[1]{#1} 
\newcommand*{\affmark}[1][*]{\textsuperscript{#1}}

\usepackage[breaklinks=true,bookmarks=false]{hyperref}

\iccvfinalcopy 

\def\iccvPaperID{5925} 

\begin{document}

\title{Photo-Realistic Monocular Gaze Redirection \\ Using Generative Adversarial Networks}

\author{%
Zhe He\affmark[1, 2],
Adrian Spurr\affmark[1],
Xucong Zhang\affmark[1],
Otmar Hilliges\affmark[1] \\
\affaddr{
\affmark[1]AIT Lab, ETH Z\"urich\\
\affmark[2]Institute of Neuroinformatics, ETH Z\"urich \& University of Z\"urich}\\
{\tt\small zhehe@student.ethz.ch, \{adrian.spurr, xucong.zhang, otmar.hilliges\}@inf.ethz.ch}
}

\maketitle

\begin{abstract}
Gaze redirection is the task of changing the gaze to a desired direction for a given monocular eye patch image.
Many applications such as videoconferencing, films, games, and generation of training data for gaze estimation require redirecting the gaze, without distorting the appearance of the area surrounding the eye and while producing photo-realistic images.
Existing methods lack the ability to generate perceptually plausible images.
In this work, we present a novel method to alleviate this problem by leveraging generative adversarial training to synthesize an eye image conditioned on a target gaze direction.
Our method ensures perceptual similarity and consistency of synthesized images to the real images. Furthermore, a gaze estimation loss is used to control the gaze direction accurately.
To attain high-quality images, we incorporate perceptual and cycle consistency losses into our architecture.
In extensive evaluations we show that the proposed method outperforms state-of-the-art approaches in terms of both image quality and redirection precision.
Finally, we show that generated images can bring significant improvement for the gaze estimation task if used to augment real training data.
\end{abstract}

\section{Introduction}
\label{sec: intro}

In the cognitive sciences it is well understood that gaze plays a crucial rule in social communication~\cite{kleinke1986gaze}, since it conveys important non-verbal cues such as emotion, intention and attention. 
Hence, many applications such as video-conferencing and movies would benefit from the ability to redirect the gaze in images to establish eye-contact with the viewer.
Furthermore, learning-based gaze estimation has recently made significant progress based on in-the-wild datasets \cite{krafka2016eye,zhang15_cvpr}. 
However, such data is difficult to acquire and datasets often only cover a restricted range of gaze angles due to the collection devices. 
A high-fidelity gaze redirection technique could be leveraged to alleviate this issue by synthesizing novel samples to augment existing datasets.

\begin{figure}[t!]
    \centering
    \includegraphics[width=\linewidth]{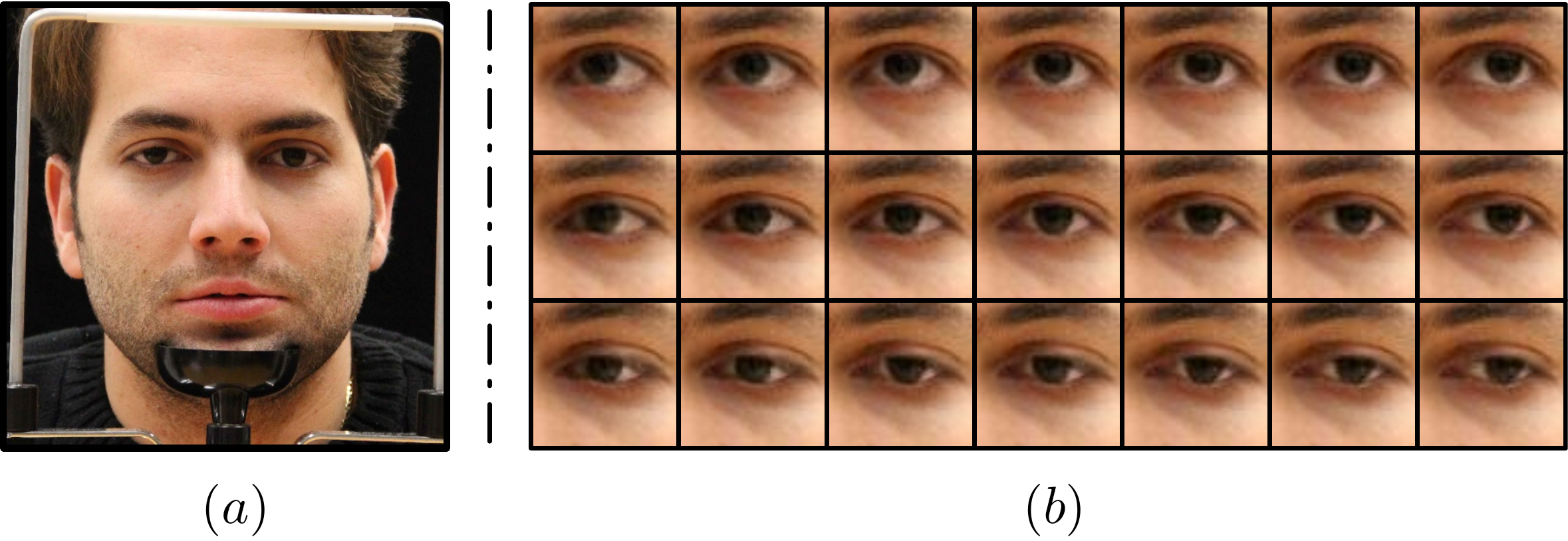}
    \caption{Gaze redirection on Columbia Gaze dataset \cite{smith2013gaze}. (a) Gaze of the subject in the source image is fully centered. (b) Sequence of eye images with different gaze directions synthesized by our method.}
    \label{fig: demo}
\end{figure}

A reliable and robust gaze redirection approach must be able to \begin{enumerate*}[label=(\alph*)] \item redirect the gaze precisely into any given direction, and \item produce photo-realistic output images which preserve shape and texture details from the input images\end{enumerate*}. 
Traditional solutions re-render the entire scene by performing 3D transformations, which requires heavy instrumentation to acquire the depth information~\cite{kuster2012gaze, yang2002eye, zhu2011eye, criminisi2003gaze}.
Recently, Ganin \etal directly rearranged the pixels of the input image to rotate the gaze direction via warping flow generated by a neural network~\cite{ganin2016deepwarp}.
However, their method fails to generate photo-realistic images for large redirection angles, especially in the presence of large dis-occlusions, such as large parts of the eyeball being covered by the eyelid in the source image.
More importantly, such warping methods cannot be perceptually plausible in terms of gaze redirection, since it minimizes pixel-wise differences between the synthesized and ground-truth images without any geometric regularization.

\begin{figure*}
\begin{center}
\includegraphics[width=.92\linewidth]{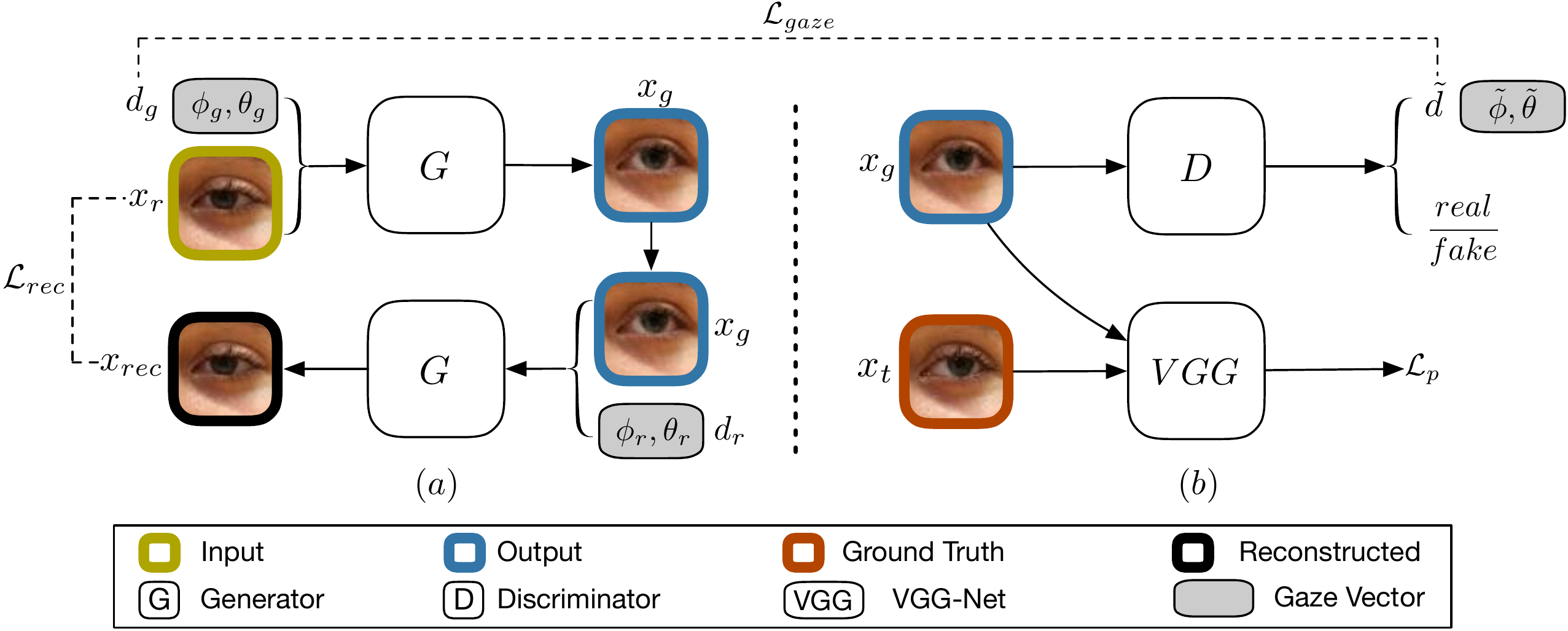}
\end{center}
   \caption{\textbf{Overview of our proposed method}. \textbf{(a)} Generator $G$ takes the original eye image $\boldsymbol{x}_r$ and target gaze direction $\boldsymbol{d}_g$ as input, and outputs the synthesized redirected eye image $\boldsymbol{x}_g$. And then $\boldsymbol{x}_g$ and the source gaze direction $\boldsymbol{d}_r$ are fed into $G$ to reconstruct the $\boldsymbol{x}_rec$.  \textbf{(b)} Discriminator $D$ is trained to discriminate real and synthesized images, and it also estimates the gaze direction to calculate the gaze estimation loss. $VGG$ takes synthesized image $\boldsymbol{x}_g$ and ground-truth image $\boldsymbol{x}_t$ to produces perceptual losses for the refinement of generated images. Please refer to Sec. \ref{sec: objectives} for details.}
\label{fig:overview}
\end{figure*}

To address the limitations of previous methods, we propose a novel gaze redirection method that builds upon generative adversarial networks (GANs)~\cite{goodfellow2014generative}. To the best of our knowledge, this is the first approach applying GANs to gaze redirection.

As shown in Fig.~\ref{fig: demo}, the proposed method can output photo-realistic eye images from a single monocular RGB image, while accurately preserving the desired gaze directions.
More specifically, we use a conditional GAN~\cite{mirza2014conditional} as backbone for our architecture shown in Fig.~\ref{fig:overview}. The generator $G$ takes a real eye image as input and generates a new synthetic eye image. Our main contribution is a novel discriminator $D$ that serves the dual purpose of i) ensuring that generated images are realistic, as is common in many GAN formulations, and ii) ensuring that the gaze direction in the output coincides with the input gaze direction which was fed to the generator. This is achieved by incorporating a gaze estimator into the discriminator network. 
Furthermore, we seek to enhance the perceptual similarity between the generated patch and its ground-truth reference. 
To this end, we utilize a perceptual loss that penalizes discrepancies between features extracted from the generated images and the ground-truth images by a separate pre-trained neural network.
Finally, to ensure that personalized features are not lost in the process of gaze redirection, we use a cycle-consistency loss that enforces consistency between the source image and the generated eye-patch. \par

We evaluate our method in quantitative experiments and via a qualitative user study. 
Furthermore, we argue that the pixel-wise difference as a metric of image quality is not suitable for the task of gaze redirection, since it does not correlate with visual perception.
To address this, we propose to use LPIPS~\cite{zhang2018unreasonable}, image blurriness and gaze estimation error as metrics for our quantitative evaluations. Providing further evidence for the high-quality of the generated images, we show in a controlled experiment that the synthetic samples can be used to augment the training data for a gaze estimation network. 
Our results show significant improvements in terms of angular gaze error compared to training with real images only. This suggests that our method can be an important tool to further enhance the accuracy attained by deep-learning based gaze estimators. \par

Our main contributions can be summarized as follows:
\begin{itemize}
    \item We propose a novel gaze redirection approach in monocular eye images. Technically this is achieved via a feature loss, gaze regularization, and adversarial training. To the best of our knowledge, it is the first GANs-based method for this task.
    \item We conduct thorough qualitative and quantitative evaluations on the gaze redirection task, showing that our method achieves state-of-the-art performance.
    \item Finally, we show the potential of leveraging gaze redirection to synthesize training data for the gaze estimation task via training data augmentation.
\end{itemize}

\section{Related Work}
\textbf{Gaze Manipulation} Approaches that redirect gaze can be divided into two groups: novel-view synthesis and monocular-gaze synthesis.

Novel-view synthesis methods~\cite{kuster2012gaze, yang2002eye, zhu2011eye, criminisi2003gaze} render a scene containing the face of a subject from a given viewpoint to mimic gazing at the camera. These methods require a depth map of the face, and then synthesize a new image of the subject with redirected gaze by performing 3D transformations.
These approaches mainly serve the purpose of correcting gaze in video conferencing, where the camera is placed at a fixed distance from the screen.
However, these methods require dedicated hardware to acquire depth. Furthermore, they alter the entire scene, which limits their applicability. \par

Monocular-gaze synthesis also aim to change the gaze within the eye region. 
Wolf \etal \cite{wolf2010eye} proposes to replace the eyes in the image with eyes from the same person while looking into a different direction~. 
Although this method retains the realism of eyes after editing, it requires collecting abundant eye images in advance. Furthermore, the movements of the eyelid are ignored in this approach. 
Recently, a number of warping-based methods have been proposed~\cite{ganin2016deepwarp, kononenko2015learning}. 
These methods use random forests or deep neural networks to learn a flow field to move pixels from the input image to the output image with the desired gaze direction.
However, such methods can not handle situations where part of the eye is occluded, since they only replace pixels with existing pixels from the original image without generating any new pixels.
Euclidean distance is commonly used as error metric in warping-based methods~\cite{ganin2016deepwarp,kononenko2015learning}.
However, this does not accurately reflect the perceptual difference between images. 
A number of approaches based on 3D modeling have been proposed~\cite{banf2009example, wood2018gazedirector}.
A 3D model is used to fit both texture and shape of the source eye patch, and then the synthesized eyeballs are superimposed on the source image. 
However, modeling methods make strong assumptions that do not hold in practice. Therefore, they can not handle images with eyeglasses and other high-variability inter-personal differences. \par

\textbf{Generative Adversarial Networks} GANs~\cite{goodfellow2014generative} have successfully been applied to many computer vision tasks, such as image super-resolution~\cite{ledig2017photo} and image compression~\cite{agustsson2018generative}, and a myriad of further variants have been proposed in recent years (e.g.,~\cite{mao2017least, arjovsky2017wasserstein, berthelot2017began, berthelot2017began,gulrajani2017improved}). GAN-based approaches have also been proposed for the task of image-to-image translation, resulting in impressive results~\cite{mirza2014conditional, isola2017image}.
However, these methods typically require paired data to train.
Zhu \etal proposed CycleGAN which functions without such requirement~\cite{zhu2017unpaired}.
Several derivatives of CycleGAN exist for various tasks~\cite{huang2018multimodal, wu2018reenactgan}.
Our method is based on the GAN model while differing from these works in two aspects.
First, we focus on a different task, namely that of gaze redirection. 
Second, we use a number of special purpose losses, including a perceptual loss between ground-truth and synthesized images and a gaze direction preservation loss for training, which we show experimentally to significantly impact the models performance. 

\section{Approach}

\subsection{Overview}
\label{sec: objectives}

Our goal is to learn a generator $G$ which can redirect the eye gaze contained in an image into any direction. Given an RGB image of an eye patch $\boldsymbol{x}_r \in \mathbb{R}^{H \times W \times 3}$ and a target gaze direction vector $\boldsymbol{d}_g = [\phi_g, \theta_g]$, where $\phi_g \in \mathbb{R}$ and $\theta_g \in \mathbb{R}$ denote the target yaw and pitch angles respectively, the task is to redirect the gaze depicted in $\boldsymbol{x}_r$ to correspond to the angles of the target vector $\boldsymbol{d}_g$, resulting in an output image $\boldsymbol{x}_g$. This output needs to satisfy two requirements. First, it needs to look real and consistent. This requires that both shape and texture of $\boldsymbol{x}_g$ are indistinguishable from those of real data. To this end, we employ a discriminator $D$ that discriminates between generated and real eye images. In order to refine the generated image more, we introduce a feature-based loss that penalizes discrepancies between generated images and ground-truth images.
Second, the eye gaze direction in $\boldsymbol{x}_g$ should look in the direction that the target gaze $\boldsymbol{d}_g$ indicates. This is accomplished via an auxiliary eye gaze estimator $D_{gaze}$ that enforces the gaze direction. Fig. \ref{fig:overview} provides the full overview of the method. We discuss the components in more detail below.


\subsection{Objectives}
Our method extends the GAN framework via integration of novel loss terms discussed below. The backbone is formed by an existing conditional GAN framework.

\noindent\textbf{Adversarial Loss} We build upon WGAN-GP \cite{gulrajani2017improved} due to its stable performance and adopt its adversarial loss to train the discriminator $D$ and generator $G$, extending $G$ to take conditional input:
\begin{equation}
\label{equation: adv loss}
    \begin{aligned}
        \mathcal{L}_{adv} = \mathbb{E}_{\boldsymbol{x}_r \sim p_{\boldsymbol{x}_r}(\boldsymbol{x})}[D_{adv}(\boldsymbol{x}_r) - D_{adv}(G(\boldsymbol{x}_r, \boldsymbol{d}_g))] + \\
        \lambda_{gp} \mathbb{E}_{\hat{\boldsymbol{x}} \sim p_{\hat{\boldsymbol{x}}}(\hat{\boldsymbol{x}})}[(\left\| \nabla_{\hat{\boldsymbol{x}}}D_{adv}(\hat{\boldsymbol{x}})\right\|_2 - 1)^2]
    \end{aligned}
\end{equation}
In Eq. \ref{equation: adv loss},  $p_{x_r}(\boldsymbol{x})$ denotes the probability distribution of real images. $D_{adv}(\boldsymbol{x})$ is the output of the discriminator. The last term is the gradient penalty, which is used to maintain the 1-Lipschitz continuity of $D_{adv}$. The hyperparameter $\lambda_{gp}$ controls the strength of gradient penalty, and we use $\lambda_{gp}=10$ in all experiments. \par

\textbf{Gaze Estimation Loss} One of our core contributions is the incorporation of an auxiliary gaze estimator $D_{gaze}$ into the GAN framework. $D_{gaze}$ is trained on real images and gaze direction pairs $(\boldsymbol{x}_r, \boldsymbol{d}_r)$ using MSE loss:
\begin{equation}
\label{equation: estimator loss d}
    \begin{aligned}
        \mathcal{L}_{gaze}^D = \mathbb{E}_{\boldsymbol{x}_r \sim p_{\boldsymbol{x}_r}(\boldsymbol{x})} \norm{\boldsymbol{d}_r - D_{gaze}(\boldsymbol{x}_r)}^2_2 \quad ,
    \end{aligned}
\end{equation}
where in practice, $D_{gaze}$ shares some layers with $D_{adv}$. 

For training $G$, the generated image $\boldsymbol{x}_g = G(\boldsymbol{x}_r, \boldsymbol{d}_g)$ is fed into the gaze estimator $D_{gaze}$. Discrepancies between the estimated gaze $D_{gaze}(\boldsymbol{x}_g)$ and the target gaze $\boldsymbol{d}_g$ are used as a loss to penalize $G$. More specifically, we add the following loss function to the training objective of $G$, keeping the weights of $D_{gaze}$ fixed:
\begin{equation}
\label{equation: estimator loss g}
    \begin{aligned}
        \mathcal{L}_{gaze}^G = \mathbb{E}_{\boldsymbol{x}_r \sim p_{\boldsymbol{x}_r}(\boldsymbol{x})} \norm{\boldsymbol{d}_g - D_{gaze}(G(\boldsymbol{x}_r, \boldsymbol{d}_g))}^2_2
    \end{aligned}
\end{equation}

\textbf{Reconstruction Loss} The above two loss terms can force the generated eye patch images to be photo-realistic, and ensure redirection of the gaze directions simultaneously. However, none of these losses ensure that personalized features, such as eyeglasses, skin tone or eyebrow are maintained during the redirection process. This is an important feature in many of the envisioned application scenarios such as video-conferencing or interactive videos. Following \cite{zhu2017unpaired} we enforce cycle consistency, penalizing bad reconstruction as follows:
\begin{equation}
\label{equation: x_rec}
    \begin{aligned}
        \boldsymbol{x}_{rec} = G(G(\boldsymbol{x}_r, \boldsymbol{d}_g), \boldsymbol{d}_r)
    \end{aligned}
\end{equation}
\begin{equation}
\label{equation: reconstruction loss}
    \begin{aligned}
        \mathcal{L}_{rec} = \mathbb{E}_{\boldsymbol{x}_r \sim p_{\boldsymbol{x}_r}(\boldsymbol{x})} \norm{\boldsymbol{x}_r - \boldsymbol{x}_{rec}}_1
    \end{aligned}
\end{equation}

Here we ask the network to first redirect the gaze to a desired direction and consecutively we generate a third image with the original gaze as target. Above loss ensures that the input and twice-encoded image are as similar as possible.

By penalizing the reconstruction discrepancies, we force the generator to maintain personalized features of the eye, which otherwise would be lost. We use the $L1$ loss, since it empirically performed better in comparison to the $L2$ loss.


\textbf{Perceptual Loss} 
In our task, human gaze only depends on pitch and yaw angles, which makes it easy to attain a ground-truth gaze images by simply asking the subject to look at the target direction. These ground-truth images can also be incorporated into the training process. One possible approach is to use Mean Squared Error (MSE) between the ground-truth images and generated images as a penalty term. However, applying a MSE loss on generated images would be too strict, as it penalizes pixel-wise discrepancies in all aspects, where minor misalignment could lead to a large MSE while humans would hardly be able to tell the differences (see Table. \ref{table: lpips}). Alternatively, we adopt the perceptual losses proposed in \cite{johnson2016perceptual} to penalize the generator $G$ for generating images which do not match ground-truth images perceptually. For this purpose, we use a VGG-16 net \cite{simonyan2014very} pre-trained on ImageNet \cite{krizhevsky2012imagenet}. \par

Let $\psi$ denote the pre-trained VGG-16 network, $\psi_j(\boldsymbol{x}) \in \mathbb{R}^{H_j \times W_j \times C_j}$ is the activation of $j$-th layer of $\psi$. Two perceptual losses, the content loss $\mathcal{L}_c$ and style loss $\mathcal{L}_s$, are defined as follows,
\begin{equation}
    \label{equation: content loss}
    \begin{aligned}
        \mathcal{L}_c = \mathbb{E}_{\boldsymbol{x}_r \sim p_{\boldsymbol{x}_r}(x)}[ \frac{1}{H_j W_j C_j} \norm{\psi_j(G(\boldsymbol{x}_r, \boldsymbol{d}_g)) - \psi_j(\boldsymbol{x}_t)}^2]
    \end{aligned}
\end{equation}
\begin{equation}
    \label{equation: style loss}
    \begin{aligned}
        \mathcal{L}_s = \mathbb{E}_{\boldsymbol{x}_r \sim p_{\boldsymbol{x}_r}(x)}[ \sum_{j=1}^{J} \norm{f_j(G(\boldsymbol{x}_r, \boldsymbol{d}_g)) - f_j(\boldsymbol{x}_t)}^2]
    \end{aligned}
\end{equation}

In Equation \ref{equation: style loss}, $\mathcal{L}_s$ is the sum of all style losses from the 1-st layer to the $J$-th layer of the VGG net. $f_j(\boldsymbol{x})$ denotes the Gram matrix, which is defined as:
\begin{equation}
    \label{equation: gram matrix}
    \begin{aligned}
        f_j(\boldsymbol{x})_{c, c'} = \frac{1}{N_j} \sum_{h}^{H_j}\sum_{w}^{W_j} \psi_j(\boldsymbol{x})_{h,w,c} \psi_j(\boldsymbol{x})_{h,w,c'}
    \end{aligned}
\end{equation}
\begin{equation}
    \label{equation: factor}
    \begin{aligned}
    N_j = H_j W_j C_j
    \end{aligned}
\end{equation}

Optimizing the content loss encourages $\boldsymbol{x}_g$ to perceptually resemble $\boldsymbol{x}_t$ in terms of overall structure and spatial relation. Meanwhile, by minimizing the style loss, the generator tries to refine the details of $\boldsymbol{x}_g$, such as color and texture, to increase the similarity to $\boldsymbol{x}_t$. The perceptual loss is the sum of content loss and style loss:
\begin{equation}
    \label{equation: perceptual loss}
    \begin{aligned}
    \mathcal{L}_{p} = \mathcal{L}_{c}  + \mathcal{L}_s
    \end{aligned}
\end{equation}

\textbf{Overall Objectives} The final training objectives consists of two parts, one for $G$ and $D$ respectively:
\begin{equation}
\label{equation: L_G}
    \begin{aligned}
        \mathcal{L}_G = -\mathcal{L}_{adv} + \lambda_{p} \mathcal{L}_p + \lambda_{gaze} \mathcal{L}_{gaze}^G + \lambda_{rec} \mathcal{L}_{rec}
    \end{aligned}
\end{equation}
\begin{equation}
\label{equation: L_D}
    \begin{aligned}
        \mathcal{L}_D = \mathcal{L}_{adv} + \lambda_{gaze} \mathcal{L}_{gaze}^D
    \end{aligned}
\end{equation}
Where $\lambda_{p}$, $\lambda_{gaze}$ and $\lambda_{rec}$ are the hyperparameters that control the contribution of each loss term. In all experiments, we set them to $\lambda_{p} = 100$, $\lambda_{gaze} = 5$, $\lambda_{rec}=50$.

\section{Implementation}

\subsection{Network Architecture}

\textbf{Generator} The generator takes an RGB eye patch image $\boldsymbol{x} \in \mathbb{R}^{H\times W\times 3}$ and a gaze direction vector $\boldsymbol{d} \in \mathbb{R}^{2}$ as input. $\boldsymbol{d}$ is expanded into $\mathbb{R}^{H\times W\times 2}$ by channel-wise duplication, such that $\boldsymbol{x}$ and $\boldsymbol{d}$ can be concatenated depth-wise. We use a modified variant of the generator architecture introduced in \cite{johnson2016perceptual}, the details of which can be found in the supplementary.

\textbf{Discriminator} We modified the last layer of the discriminator architecture of WGAN-GP \cite{gulrajani2017improved} to have two output branches: one performs real/fake discrimination and another one outputs gaze estimates respectively.

\textbf{VGG-16} We use the standard architecture of VGG-16 introduced in~\cite{simonyan2014very}. We use the activation of the 5th layer to produce the content loss, and the 1st to 4th layers to produce the style loss.

\subsection{Training Details}
For all the following experiments, we use Adam \cite{kingma2014adam} optimizer with $\beta_1 = 0.5$, $\beta_2=0.999$.
Our model is trained for 300 epochs with batch size 32. 
The learning rate is set to 0.0002 for the first 150 epochs, and linearly decays to 0 during the next 150 epochs. 
For every update of the generator, we update the discriminator five times. 
The training process takes about 16 hours on a single NVIDIA\textsuperscript{\textregistered} 1080Ti GPU.

\section{Experiments}

In this section, we detail the quantitative and qualitative experiments that were conducted to evaluate our approach. 

\subsection{Metrics}
As mentioned before (see Sec. \ref{sec: intro}), gaze redirection models are required to be precise in redirecting and to produce photo-realistic and consistent images. 
Correspondingly, the evaluation metrics need to be able to assess these aspects. 
In previous work of monocular gaze manipulation~\cite{ganin2016deepwarp, wood2018gazedirector}, the mean squared error (MSE) was used as the metric to measure the similarity between the generated eye images and ground-truth eye images. 
This was used as a quantitative measure of performance.
However, we argue that MSE is not the ideal metric for this task, as has been observed previously in related work~\cite{wang2009mean}.
To illustrate the issue, we created three types of image degradations compared to the ground-truth as shown in Table. \ref{table: lpips}. 
Qualitatively, Table. \ref{table: lpips} d) is the most similar to the ground-truth Table. \ref{table: lpips} a). 
However, when calculating the MSE, we see that this does not correlate well with one's qualitative judgment.

\begin{table}[t]
\begin{center}
\small
\begin{tabular}{|c|c|c|c|}
    \hline
    (a) Eye patch & \small(b) Blurred & \small(c) Noisy & \small(d) Shifted \\
    \hline
    \raisebox{-0.3\height}{\includegraphics[width=.18\linewidth]{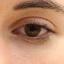}} &
    \raisebox{-0.3\height}{\includegraphics[width=.18\linewidth]{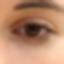}} &
    \raisebox{-0.3\height}{\includegraphics[width=.18\linewidth]{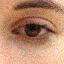}} &
    \raisebox{-0.3\height}{\includegraphics[width=.18\linewidth]{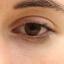}} \\
    \hline \hline
    \textbf{MSE} & \textbf{69.57} & 155.36 & 176.06 \\
    \hline
    \textbf{LPIPS} & 0.122 & 0.106 & \textbf{0.016} \\
    \hline
\end{tabular}
\end{center}
\caption{Examples of image degradations. (\textbf{a}) Eye patch from training set. (\textbf{b}) Blurred with Gaussian filter. (\textbf{c}) With random Gaussian noise. (\textbf{d}) Shifted up by one pixel.}
\label{table: lpips}
\end{table}

Instead, we propose to use the following three error metrics: LPIPS score, image blurriness and gaze estimation error.

\textbf{LPIPS Score.} We use the Learned Perceptual Image Patch Similarity (LPIPS)~\cite{zhang2018unreasonable} metric to evaluate the visual quality of the generated gaze images.
Different from traditional metrics, LPIPS is based on deep networks and aims to resemble human perception in image evaluation tasks. 
The LPIPS score is given as follows:
\begin{equation}
\label{equation: lpips}
    \begin{aligned}
        d(\boldsymbol{x}, \boldsymbol{x}_0) = \sum_l \frac{1}{H_lW_l} \sum_{h,w} \norm{\boldsymbol{w}_l \odot (\boldsymbol{\hat{y}}_{hw}^l - \boldsymbol{\hat{y}}_{0hw}^l)}_2^2
    \end{aligned}
\end{equation}
Where $d(\boldsymbol{x}, \boldsymbol{x}_0)$ denotes the LPIPS score between the images $\boldsymbol{x} \in \mathbb{R}^{H \times W \times 3}$ and $\boldsymbol{x}_0 \in \mathbb{R}^{H \times W \times 3}$. The variables $\boldsymbol{\hat{y}}^l \in \mathbb{R}^{H_l \times W_l \times C_l}$ and $\boldsymbol{\hat{y}}_{0}^l \in \mathbb{R}^{H_l \times W_l \times C_l}$ are the channel-wise unit-normalized activation from the $l$-th layer of the backbone network and
$\boldsymbol{w}_l \in \mathbb{R}^{C_l}$ are the trainable weights used for scaling the activations. In our work, we use the pre-trained Alex-Net \cite{krizhevsky2012imagenet} as a backbone,

When calculating LPIPS on the previous examples in Table. \ref{table: lpips}, we see that the scores agree more with human evaluation.

\textbf{Image Blurriness (IB).} To measure the blurriness of a generated gaze image, we use a Laplace filter $k$ and perform convolution on the grayscale gaze image $\boldsymbol{x}_{gray}$. Image blurriness can be acquired by calculating the reciprocal variance of the filtered image as shown in the following equations:


\begin{equation}
    \label{equation: blurriness}
    \begin{aligned}
    k=
  \begin{bmatrix}
    0 & 1 & 0 \\
    1 & -4 & 1 \\
    0 & 1 & 1
  \end{bmatrix},
        \mathrm{IB} = \frac{1}{\mathrm{Var}[k * \boldsymbol{x}_{gray}]}
    \end{aligned}.
\end{equation}

\textbf{Gaze Estimation Error.} For the assessment of gaze redirection accuracy, we employ a state-of-the-art gaze estimator proposed by Park \etal \cite{park2018deep} which was pre-trained on MPIIGaze \cite{zhang15_cvpr}. 
The estimator predicts the gaze direction of the generated gaze images. 
The angular error $\delta$ between the target gaze direction $\boldsymbol{d}_g$ and the predicted gaze direction $\boldsymbol{\hat{d}}$ is used as the gaze estimation error. 
To attain $\delta$, the yaw and pitch angles $(\phi, \theta)$ need to be converted into three-dimensional Cartesian coordinates first:
\begin{equation}
    \label{equation: euler2cart}
        \boldsymbol{v} = T(\boldsymbol{d}) = [\cos \phi \cos \theta, -\sin \phi, \cos \phi \sin \theta].
\end{equation}

where $T(.)$ denotes the mapping between two coordinate systems. Then, $\delta$ can be obtained via the following calculations:
\begin{equation}
    \label{equation: v}
        \boldsymbol{v}_g = T(\boldsymbol{d}_g), \boldsymbol{\hat{v}} = T(\boldsymbol{\hat{d}})
\end{equation}
\begin{equation}
    \label{equation: delta}
        \delta= \arccos \frac{\boldsymbol{v}_g^\mathsf{T} \cdot \boldsymbol{\hat{v}}}{\norm{\boldsymbol{v}_g} \cdot \norm{\boldsymbol{\hat{v}}}}.
\end{equation}

\subsection{Dataset}
\label{sec: dataset}
We used the Columbia Gaze dataset \cite{smith2013gaze} for the evaluations, which is a high-resolution, publicly available human gaze dataset collected from 56 subjects.
The head poses of each subject are discrete values in the set [$-30^\circ, -15^\circ, 0^\circ, 15^\circ, 30^\circ$]. 
For each head pose, there are 21 gaze directions, which are the combinations of three pitch angles [$-10^\circ, 0^\circ, 10^\circ$], and seven yaw angles [$-15^\circ, -10^\circ, -5^\circ, 0^\circ, 5^\circ, 10^\circ, 15^\circ$].
Here, we only used the images with frontal faces, i.e. $0^\circ$ head pose. Results on non-frontal faces are provided in the supplementary.
We split the data into train and test set. 
The former set includes 50 subjects whereas the latter contains 6 subjects. 
We first run face alignment with \texttt{dlib} \cite{dlib09} by parsing the face with 68 facial landmark points. 
After that, a minimal enclosed circle with center $(x, y)$ and radius $R$ was extracted from the 6 landmark points of each eye.
The cropping region of the eye patch is set as a square box with center $(x, y)$ and side length $3.4R$. 
We flipped the right eye images horizontally to align with the left eye images. All eye patch images were resized to 64 $\times$ 64.
Both the pixel values of images and gaze directions were normalized into the range $[-1.0, 1.0]$. 
Other publicly available gaze datasets, such as MPIIGaze \cite{zhang15_cvpr} or EYEDIAP \cite{FunesMora_ETRA_2014} only provide low-resolution images and would therefore introduce a bias towards low quality images. Therefore, these datasets were not suitable for our task.
\subsection{Evaluation Protocol}

We tested each model on the 6 subjects contained in the test set, which includes 252 eye patch images. 
For each image, we redirected the gaze into 20 gaze directions separately, excluding the gaze direction of the current image.
Intuitively, it would be harder for the model to redirect the gaze if the target gaze direction is significantly different from the original gaze direction.
Therefore, we defined the correction angle $\gamma$ to indicate the angular difference between original and target gaze directions. 
It is calculated as follows:
\begin{equation}
\label{equation: v_g}
    \begin{aligned}
        \boldsymbol{v}_g = T(\boldsymbol{d}_g), \boldsymbol{v}_r = T(\boldsymbol{d}_r)
    \end{aligned}
\end{equation}
\begin{equation}
    \label{equation: gaze estimation error}
        \gamma = \arccos \frac{\boldsymbol{v}_g^\mathsf{T} \cdot \boldsymbol{v}_r}{\norm{\boldsymbol{v}_g} \cdot \norm{\boldsymbol{v}_r}}
\end{equation}
Where $T(.)$ is the aforementioned mapping in Eq.~\ref{equation: euler2cart}.

\subsection{Comparison to State-of-The-Art}

\textbf{Baseline Model} We adopt DeepWarp \cite{ganin2016deepwarp} as our baseline model. 
The original implementation uses 7 eye landmarks as input, including the pupil center.
However, detecting the pupil center is very challenging task.
Therefore we only used 6 landmarks as the input to DeepWarp. 
Unfortunately, evaluating the more recent work GazeDirector \cite{wood2018gazedirector} with the proposed error metric is not possible, since their implementation is not available.
Therefore, we did not compare GazeDirector in our paper.

\begin{figure}[t]
\begin{center}

\includegraphics[width=.85\linewidth]{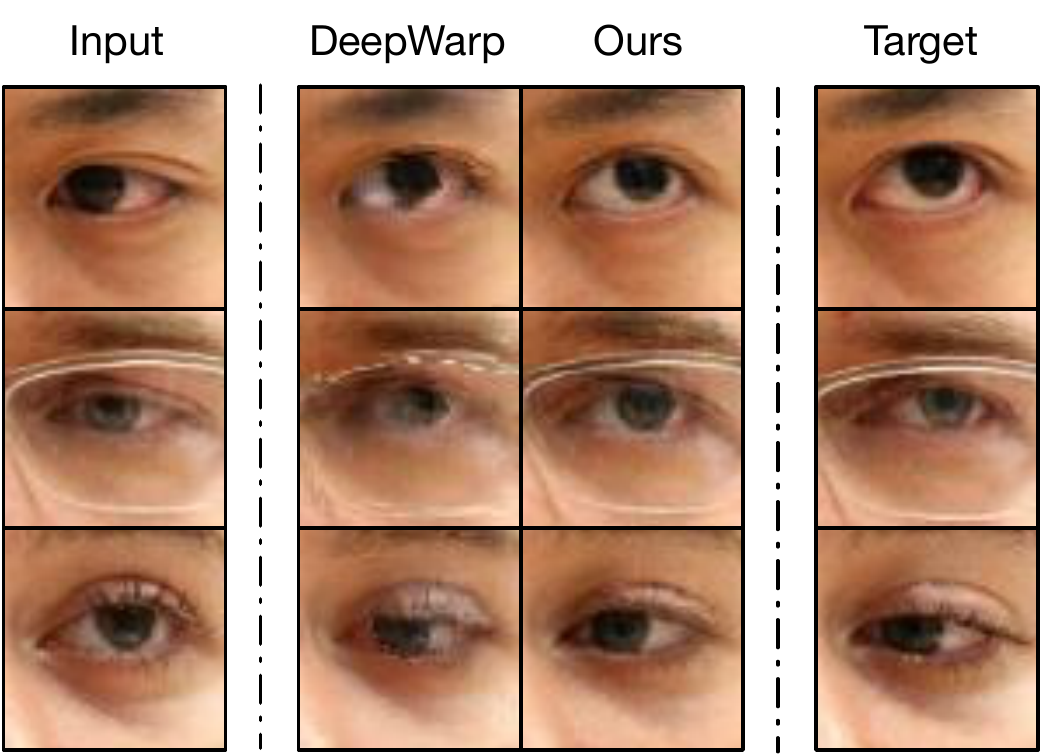}
\end{center}
   \caption{Gaze redirection comparison.}
\label{fig: comparison qualitative}
\end{figure}

\begin{figure}[t]
\begin{center}
\includegraphics[width=0.85\linewidth]{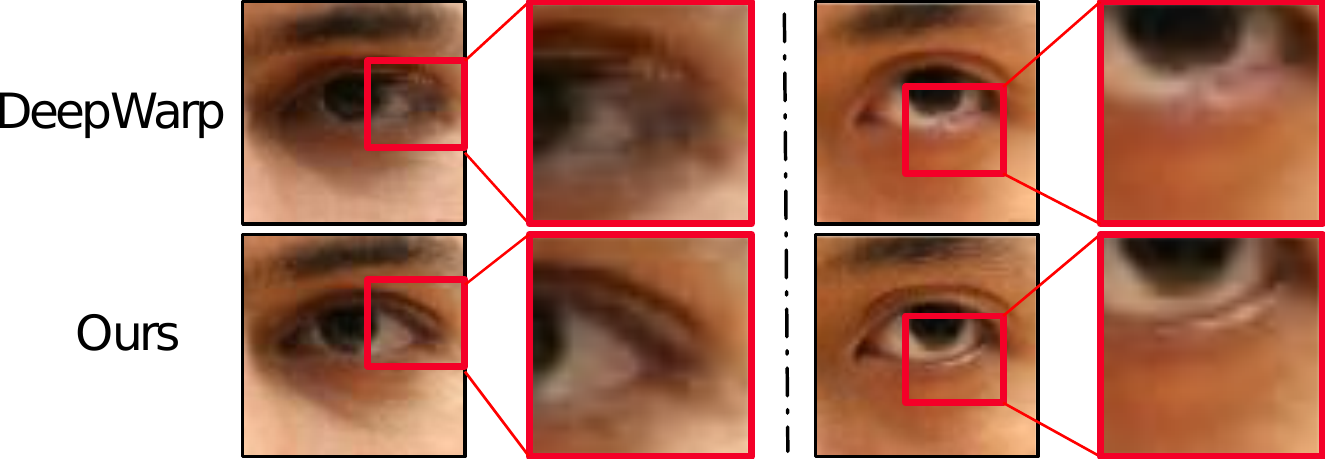}
\end{center}
   \caption{Zoom-in detail comparison.}
\label{fig: comparison qualitative details}
\end{figure}

\textbf{Qualitative Evaluation} Fig.~\ref{fig: comparison qualitative} and Fig.~\ref{fig: comparison qualitative details} show the generated gaze images examples. 
Although both methods are capable of redirecting the gaze, we observe that the generated images of DeepWarp have several obvious defects. 
First, textures such as skin and eyebrows are more blurry. 
Second, the shapes of certain parts, such as the edges of eyelid (see Fig.~\ref{fig: comparison qualitative details}), iris and eyeglasses (see Fig.~\ref{fig: comparison qualitative}), are distorted. 
In contrast, the generated gaze images of our proposed method are more faithful to the input images. 

\begin{figure}[ht]
\begin{center}
    \includegraphics[width=0.88\linewidth]{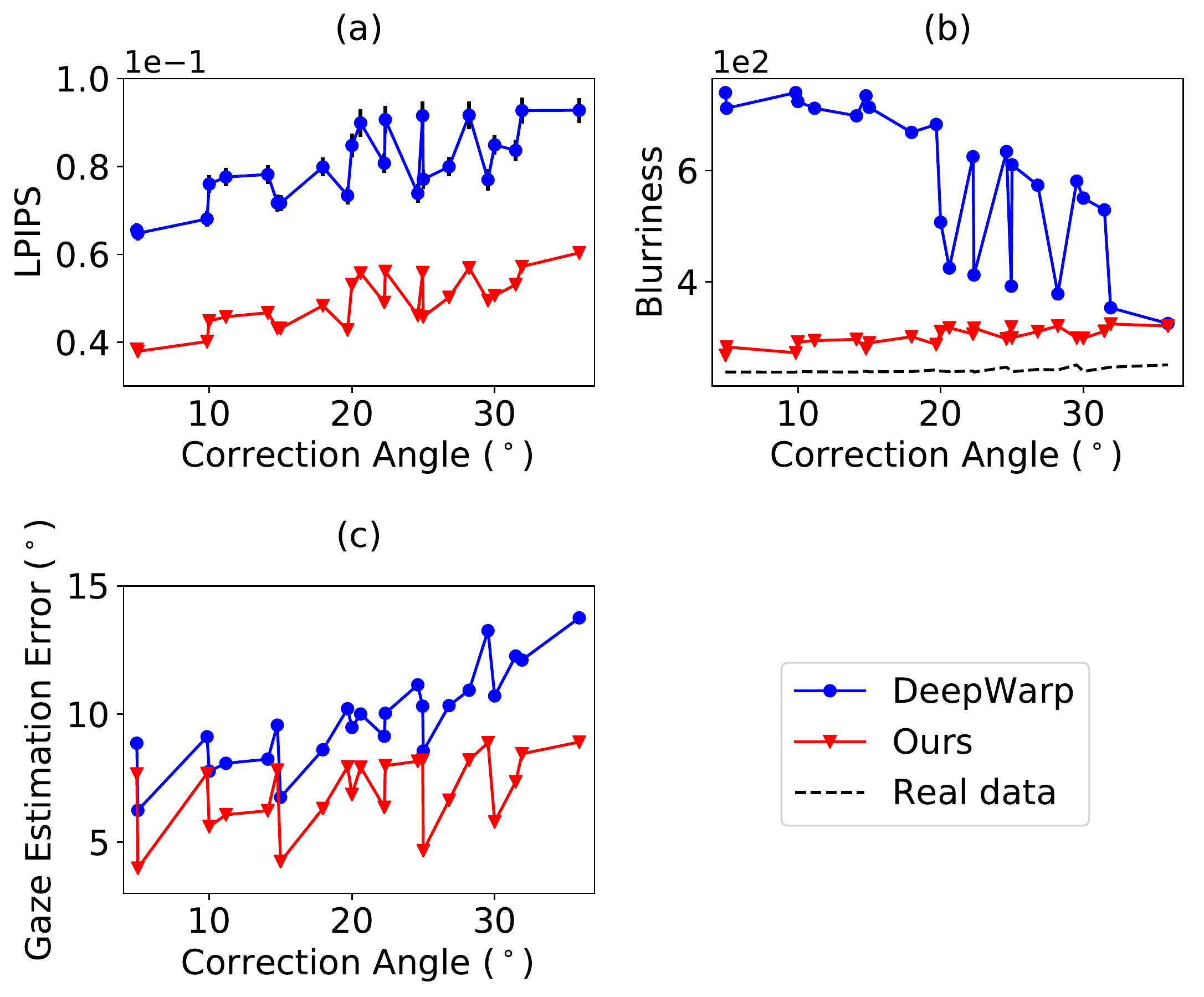}
\end{center}
   \caption{Quantitative evaluation results of DeepWarp and our method. (a) LPIPS score (lower is better). (b) Image blurriness (lower is better). (c) Gaze estimation error (lower is better)}
\label{fig: comparison}
\end{figure}

\textbf{Quantitative Evaluation} Fig.~\ref{fig: comparison}a plots the LPIPS scores of DeepWarp and our method. 
The range of correction angle is [4.9$^\circ$, 35.9$^\circ$]. 
From the figure we can see that our method achieves the lower LPIPS score than DeepWarp at every correction angle, which indicates that our method can generate gaze images that are perceptually more similar to the ground-truth images.
This observation is consistent with the qualitative evaluation (Fig.~\ref{fig: comparison qualitative}  and Fig.~\ref{fig: comparison qualitative details}). \par

Fig.~\ref{fig: comparison}b plots the blurriness of the produced images. 
Our method outperforms the related work by a large margin, being closer to the blurriness observed in real images. \par
Fig.~\ref{fig: comparison}c presents the results of gaze estimation error. 
The error of our method is much lower than DeepWarp, which indicates that our method can redirect the gaze with a higher precision. 

\begin{figure*}[t]
\begin{center}
    \includegraphics[width=.74\linewidth]{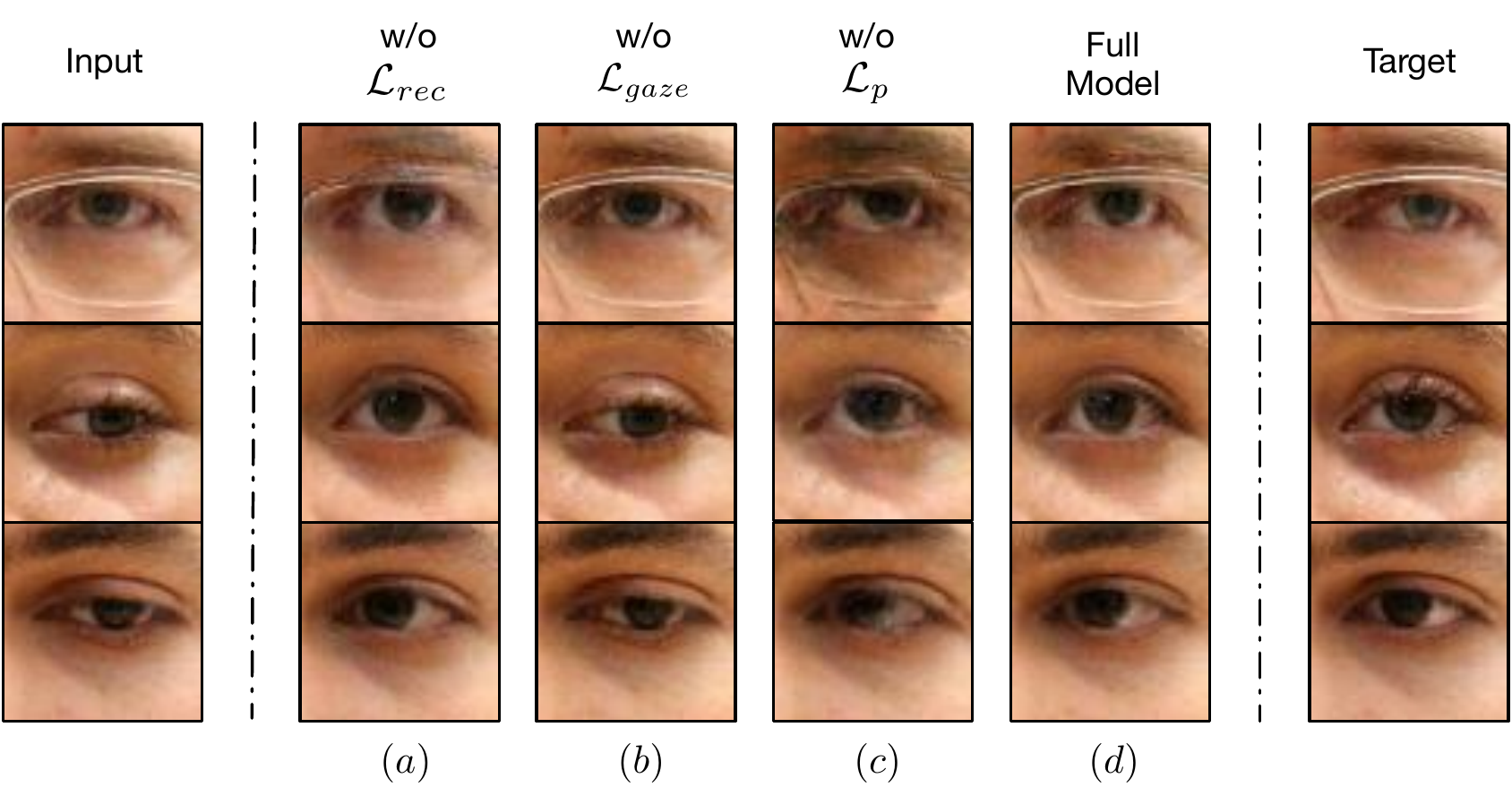}
\end{center}
   \caption{Gaze redirection results of (a) model without $\mathcal{L}_{rec}$ and (b) model without $\mathcal{L}_{gaze}$, (c) model without $\mathcal{L}_{p}$ and (d) full model.}
\label{fig: ablation qualitative}
\end{figure*}

\textbf{User Study} In addition, we conducted a user study to compare the performance of DeepWarp \cite{ganin2016deepwarp} and our method. 
As the overall range of correction angle is $[4.9^\circ, 35.9^\circ]$, we split the generated gaze images into three groups: $[4.9^\circ, 15.0^\circ]$, $(15.0^\circ, 25.0^\circ]$, $(25.0^\circ, 35.9^\circ]$, which represent the difficulty of gaze redirection from easy to hard. 
In each group, we randomly choose 19 pairs of images generated by both methods with the same input image and gaze direction. Two images in a pair were shown side by side to the user without any further information. 
The task for the users is to pick the gaze image that looks more realistic than the other. \par

In total, we have 16 users participated in our study.
Table \ref{tab: user study} shows the results of the user study. 
We can see that our method outperforms DeepWarp with a significant margin. 
The results of quantitative evaluations shown in Fig.~\ref{fig: comparison} are consistent with the user assessment, which demonstrates the the metrics we used are effective in the evaluation of the gaze redirection task.

\begin{table}[ht]
\begin{center}
\small
\begin{tabular}{|c|c|c|}
\hline
Group & DeepWarp \cite{ganin2016deepwarp} & Ours\\
\hline\hline
$[4.9^\circ, 15.0^\circ]$ & 21.9\% & \textbf{78.1\%} \\
$(15.0^\circ, 25.0^\circ]$ & 9.0\% & \textbf{91.0\%} \\
$(25.0^\circ, 35.9^\circ]$ & 13.4\% & \textbf{86.6\%} \\
\hline
\end{tabular}
\end{center}
\caption{Voting results of user study, comparing DeepWarp with our method. Each row sums up to 100 \%.}
\label{tab: user study}
\end{table}


\begin{figure}
\begin{center}
    \includegraphics[width=0.88\linewidth]{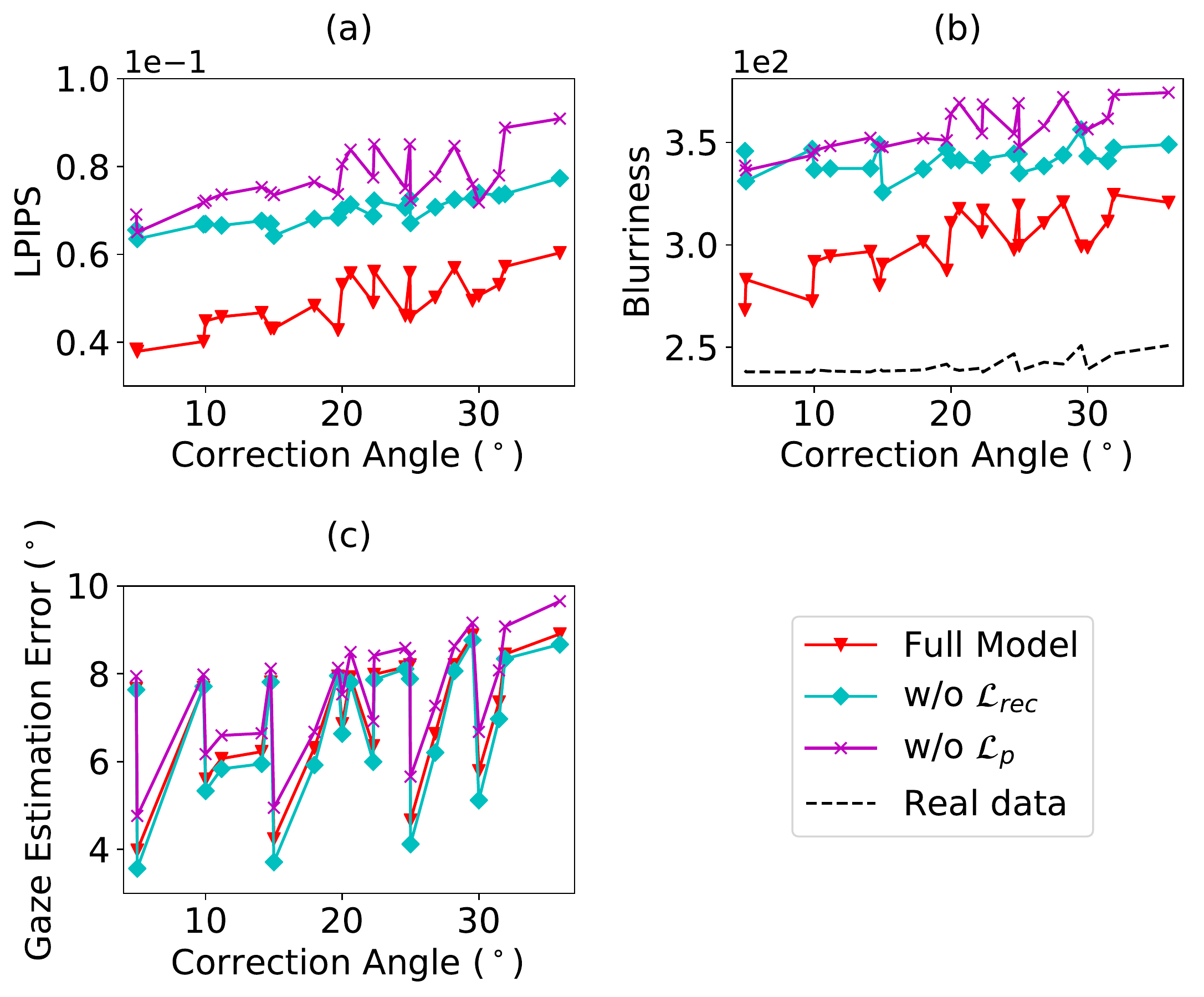}
\end{center}
   \caption{Quantitative evaluation results of our full model, model without $\mathcal{L}_{rec}$ and model without $\mathcal{L}_{p}$. (a) LPIPS score (lower is better). (b) Image blurriness (lower is better). (c) Gaze estimation error (lower is better)}
\label{fig: ablation}
\end{figure}
\subsection{Ablation Study}
\label{sec: ablation}
To understand the effect of each component of our proposed model, we performed an ablation study. 
As mentioned in Sec.~\ref{sec: objectives}, besides the adversarial loss, there are three other loss terms: gaze estimation loss, reconstruction loss and perceptual loss. 
We trained a model for each one of these additional loss terms, where one of the terms was removed from the total loss.

\textbf{Qualitative Results} We show the results in Fig.~\ref{fig: ablation qualitative}.
As can be seen from the second column of Fig.~\ref{fig: ablation qualitative}, the model is not able to maintain features from input images without $\mathcal{L}_{rec}$. 
The most significant example is in the first row, where the model without $\mathcal{L}_{rec}$ does not preserve the rim of the eyeglasses. \par
When discarding $\mathcal{L}_{gaze}$, it can be observed that the model fails to redirect the gaze entirely. 
Therefore, we will not further consider models not using $\mathcal{L}_{gaze}$ in the following quantitative evaluations. \par
Removal of $\mathcal{L}_{p}$ causes reduced image quality as can be visually verified in the generated images. These show artifacts such as distortion of eyelid shape, iris and eyeglasses (Fig.~\ref{fig: ablation qualitative}). \par

\textbf{Quantitative Results}
Fig~\ref{fig: ablation}a shows the LPIPS scores of the full model, a model without $\mathcal{L}_{rec}$ and a model without $\mathcal{L}_{p}$. It is clear that the LPIPS score increases without either of $\mathcal{L}_{rec}$ or $\mathcal{L}_{p}$, which indicates that both terms are essential for improving the visual quality of redirected gaze images. \par

The blurriness scores shown in Fig.~\ref{fig: ablation}b are also consistent with what has been observed in qualitative results, where the full model produces the sharpest images. \par

Fig.~\ref{fig: ablation}c presents the gaze estimation error. Removal of either $\mathcal{L}_{rec}$ or $\mathcal{L}_{p}$ does not significantly worsen the gaze estimation error, since the precision of redirection is mainly controlled by $\mathcal{L}_{gaze}$.

\subsection{Augmenting Gaze Data}

Lastly, we investigated the feasibility of leveraging our method for the purpose of data augmentation for eye gaze estimation tasks. 
This is motivated by the rapid progress in deep-learning based gaze estimation (e.g., \cite{zhang15_cvpr,park2018deep}). 
While appearance-based gaze estimation techniques that use Convolutional Neural Networks (CNN) have significantly surpassed classical ones \cite{zhang15_cvpr} in in-the-wild settings, there still remains a significant gap towards applicability in high-accuracy domains.
The currently lowest reported person-independent error of $4.3^\circ$ \cite{Fischer2018ECCV} is roughly equivalent to 4.7cm at a distance of 60cm. 
One reason for this relatively high error is the lack of sufficient training data. 
In particular, it is known that many datasets only cover a relatively small range of gaze angles due to hardware limitations.
Therefore we propose to leverage our model for augmenting existing datasets, in order to expand the range of gaze directions and leading to better gaze estimation performance. To the best of our knowledge, this is the first time that potential of gaze redirecting models to improve gaze estimation models have been explored.~\footnote{As of submission. Since publication we have become aware of Yu \etal~\cite{yu2019improving} performing the same task even earlier.}.

To assess the applicability of our method in this setting, we performed a proof-of-concept experiment indicating that our technique can fill in unseen gaze angles. 
First, we constructed two datasets. \par

The \textbf{raw dataset} contains all the eye images with $10^\circ$ pitch angles from the Columbia Gaze Dataset \cite{smith2013gaze}. \par

The \textbf{augmented dataset} contains the images from the raw dataset. 
Furthermore, we took the images of the 6 testing subjects (see Sec. \ref{sec: dataset}), and used them to synthesize new gaze images with pitch angles $-10^\circ$ and $0^\circ$ respectively. \par

We trained two gaze estimators on the raw and augmented datasets respectively.
Both estimators were constructed by the same VGG-16 \cite{simonyan2014very} architecture. Since augmented dataset contains more images, we trained the corresponding estimator for less epochs. Implementation details can be found in the supplementary.\par

To test the estimators, we used two test sets. (1) \textbf{Columbia Gaze.} Since the eye images in Columbia Gaze dataset with pitch angles $-10^\circ$ and $0^\circ$ of the 50 training subjects (see Sec. \ref{sec: dataset}) have not been seen by the gaze estimators, we use these images as our test set without leaking information. (2) \textbf{MPIIGaze.} For cross-dataset evaluation, we take the test set of MPIIGaze \cite{zhang15_cvpr}, where the pitch angles are in the range [$-20^\circ$, $1.5^\circ$].\par
\begin{table}[ht]
\begin{center}
\small
\begin{tabular}{|c|c|c|}
\hline
Dataset & Raw & Augmented\\
\hline\hline
Columbia & $14.3^\circ$ & $\mathbf{6.9}^\circ$ \\
MPIIGaze & $20.2^\circ$ &  $\mathbf{14.0}^\circ$ \\
\hline
\end{tabular}
\end{center}
\caption{Gaze estimation errors. Column name is the training set, while row name is the testing set.}
\label{tab: augmentation}
\end{table}

\paragraph{Results} As shown in Table \ref{tab: augmentation}, the gaze estimator trained on the augmented dataset always performs better than the gaze estimator trained on the raw dataset. 
Intuitively, since the raw dataset only contains images with positive pitch angles, the trained estimator is expected to generalize poorly on the test set, where most samples have different pitch angles.
In contrast, the augmented images aid the estimator in generalizing better to unseen angles, improving the test set performance.

\section{Conclusion}
In this paper, we propose a novel monocular gaze redirection method leveraging generative adversarial networks.
The proposed method can generate photo-realistic eye images while preserving the desired gaze direction.
In order to further refine the generated images, we incorporate a perceptual loss into the adversarial training and include a cycle-consistent loss to preserve personalized features.
Extensive evaluations show that our approach outperforms previous state-of-the-art methods in terms of both image quality and redirection precision.
Finally, we show that our gaze redirection method can benefit gaze estimation tasks by generating additional training data with controlled gaze directions.

\section{Acknowledgement}
We thank the NVIDIA Corporation for the donation of GPUs used in this work.

{\small
\bibliographystyle{ieee}
\bibliography{egbib}
}

\newpage

\clearpage
\section{Appendix}
\subsection{Network Architecture}
In this section, we provide the details of network architecture discussed in Sec.~{4.1}.
\subsubsection{Abbreviations}

\textbf{Conv($k \times k$, $s$, $p$)}: A convolutional layer with kernel size $k \times k$, stride size $s$ and padding size $p$. Zero padding is used in all convolutional layers.
\textbf{IN}: An instance normalization layer.
\textbf{ReLU}: A ReLU activation layer.
\textbf{LReLU}: A Leaky ReLU activation layer. Slope of the activation function at $x < 0$ is set to 0.01.
\textbf{Tanh}: A tanh activation layer.
\textbf{DeConv($k \times k$, $s$, $p$)}: A transposed convolutional layer with kernel size $k \times k$, stride size $s$ and padding size $p$. Zero padding is used in all transposed convolutional layers.
\textbf{Res}($k \times k$, $s$, $p$, IN, ReLU): A residual layer which builds upon Conv($k \times k$, $s$, $p$), IN and ReLU layers.

\subsubsection{Generator}
\begin{table}[h]
    \centering
        \begin{tabular}{|c|c|}
            \hline
            \textbf{Layers} & \textbf{Output} \\
            \hline
            Conv(7x7, 1, 3)--IN--ReLU  & (64, 64, 64)\\
            \hline
            Conv(4x4, 2, 1)--IN--ReLU  & (32, 32, 128)\\
            \hline
            Conv(4x4, 2, 1)--IN--ReLU  & (16, 16, 256)\\
            \hline
            Res(3x3, 1, 1, IN, ReLU)  & (16, 16, 256)\\
            \hline
            Res(3x3, 1, 1, IN, ReLU)  & (16, 16, 256)\\
            \hline
            Res(3x3, 1, 1, IN, ReLU)  & (16, 16, 256)\\
            \hline
            Res(3x3, 1, 1, IN, ReLU)  & (16, 16, 256)\\
            \hline
            Res(3x3, 1, 1, IN, ReLU)  & (16, 16, 256)\\
            \hline
            Res(3x3, 1, 1, IN, ReLU)  & (16, 16, 256)\\
            
            \hline
            DeConv(4x4, 2, 1)--IN--ReLU   & (32, 32, 128)\\
            \hline
            DeConv(4x4, 2, 1)--IN--ReLU   & (64, 64, 64)\\
            
            \hline
            Conv(7x7, 1, 3)--Tanh  & (64, 64, 3)\\
            \hline
        \end{tabular}
    \caption{Generator Architecture}
    \label{tab: generator}
\end{table}

\subsubsection{Discriminator}
\begin{table}[h!]
    \centering
        \begin{tabular}{|c|c|}
            \hline
            \textbf{Layers} & \textbf{Output} \\
            \hline
            Conv(4x4, 2, 1)--LReLU  & (32, 32, 64)\\
            \hline
            Conv(4x4, 2, 1)--LReLU  & (16, 16, 128)\\
            \hline
            Conv(4x4, 2, 1)--LReLU  & (8, 8, 256)\\
            \hline
            Conv(4x4, 2, 1)--LReLU  & (4, 4, 512)\\
            \hline
            Conv(4x4, 2, 1)--LReLU  & (2, 2, 1024)\\
            \hline
        \end{tabular}
    \caption{Backbone Network of Discriminator}
    \label{tab: backbone}
\end{table}

\begin{table}[!ht]
    \centering
        \begin{tabular}{|c|c|}
            \hline
            \textbf{Layers} & \textbf{Output} \\
            \hline
            Backbone  & (2, 2, 1024)\\
            \hline
            Conv(2x2, 1, 1)  & (3, 3, 1)\\
            \hline
        \end{tabular}
    \caption{Discriminator Architecture}
    \label{tab: discriminator}
\end{table}

\begin{table}[!ht]
    \centering
        \begin{tabular}{|c|c|}
            \hline
            \textbf{Layers} & \textbf{Output} \\
            \hline
            Backbone  & (2, 2, 1024)\\
            \hline
            Conv(2x2, 1, 0)  & (1, 1, 2)\\
            \hline
        \end{tabular}
    \caption{Gaze Estimator Architecture}
    \label{tab: gaze estimator}
\end{table}

\subsection{Implementation}
Code for training and testing our model is available online (\href{https://github.com/HzDmS/gaze_redirection}{https://github.com/HzDmS/gaze\_redirection}).

\subsection{Training Details of Gaze Estimators}
Training details of the gaze estimators used in Sec.~{5.6} are provided in this section. We used the Adam optimizer with learning rate 0.00005, $\beta_1=0.5$, and $\beta_2=0.999$. Batch size was set to 32. For the training on the raw dataset, the gaze estimator was trained for 200 epochs. For the training on the augmented dataset, the gaze estimator was trained for 100 epochs.

\subsection{Results on Non-frontal Faces}

We conducted an additional experiment on non-frontal head poses and compared them with the frontal head pose.
We used the same settings as introduced in \mbox{Sec. 4.2} and \mbox{Sec. 5.2} (in our paper).
Samples which could not be successfully parsed with \texttt{dlib} \cite{dlib09} were not included in the training and test datasets.
Note this process removed some samples with extreme head poses.

\begin{figure}[ht]
\begin{center}
   \includegraphics[width=.96\linewidth]{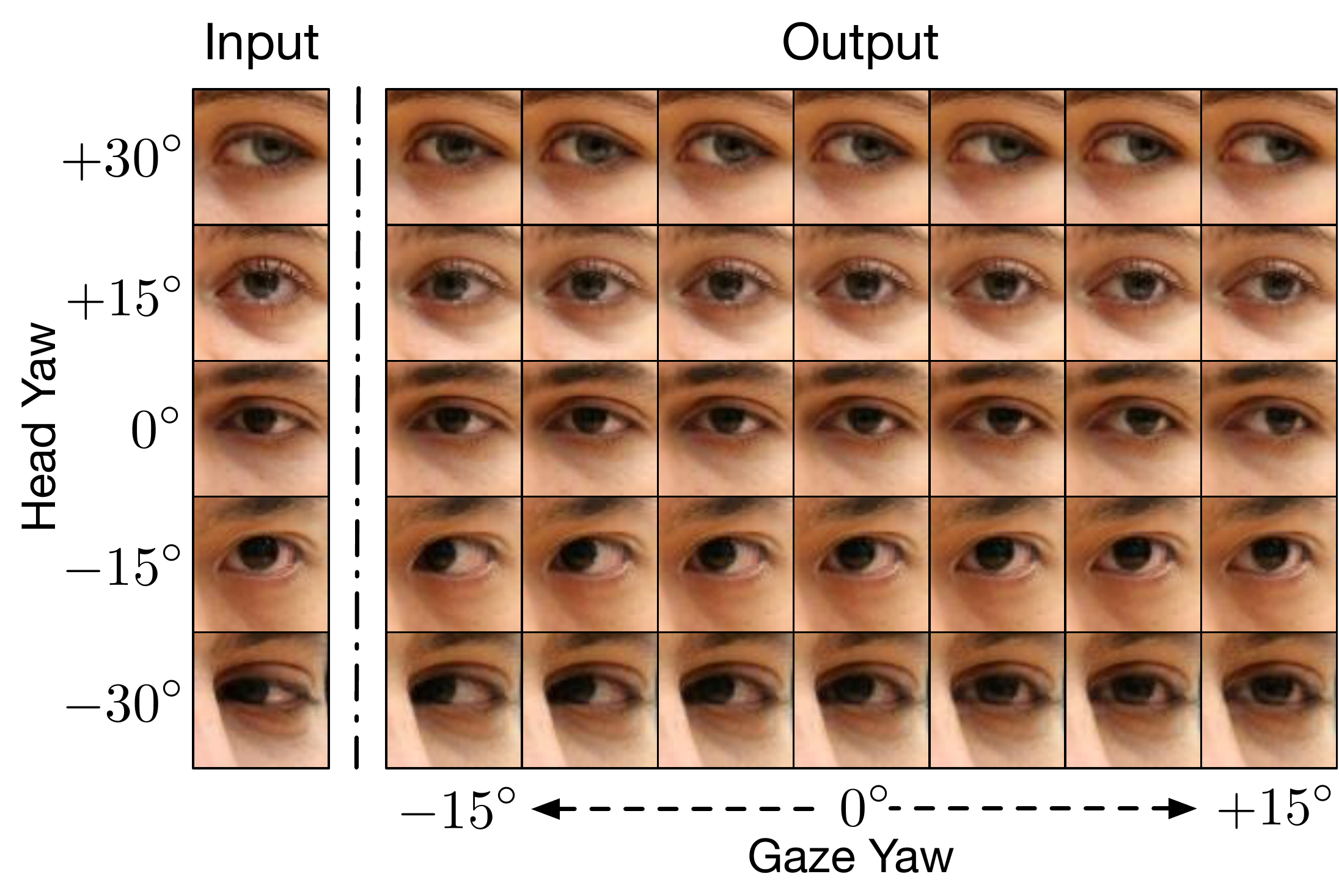}
\end{center}
   \caption{Gaze redirection results on images with different head poses. In the output images, the gaze pitch is equal to $0^\circ$}
\label{fig:qualitative}
\end{figure}

Fig. \ref{fig:qualitative} shows redirected eye-images (with $0^\circ$ output gaze pitch) using input images with varying head-poses. 
The method produces high-quality results on these inputs.

\begin{figure}
\begin{center}
   \includegraphics[width=\linewidth]{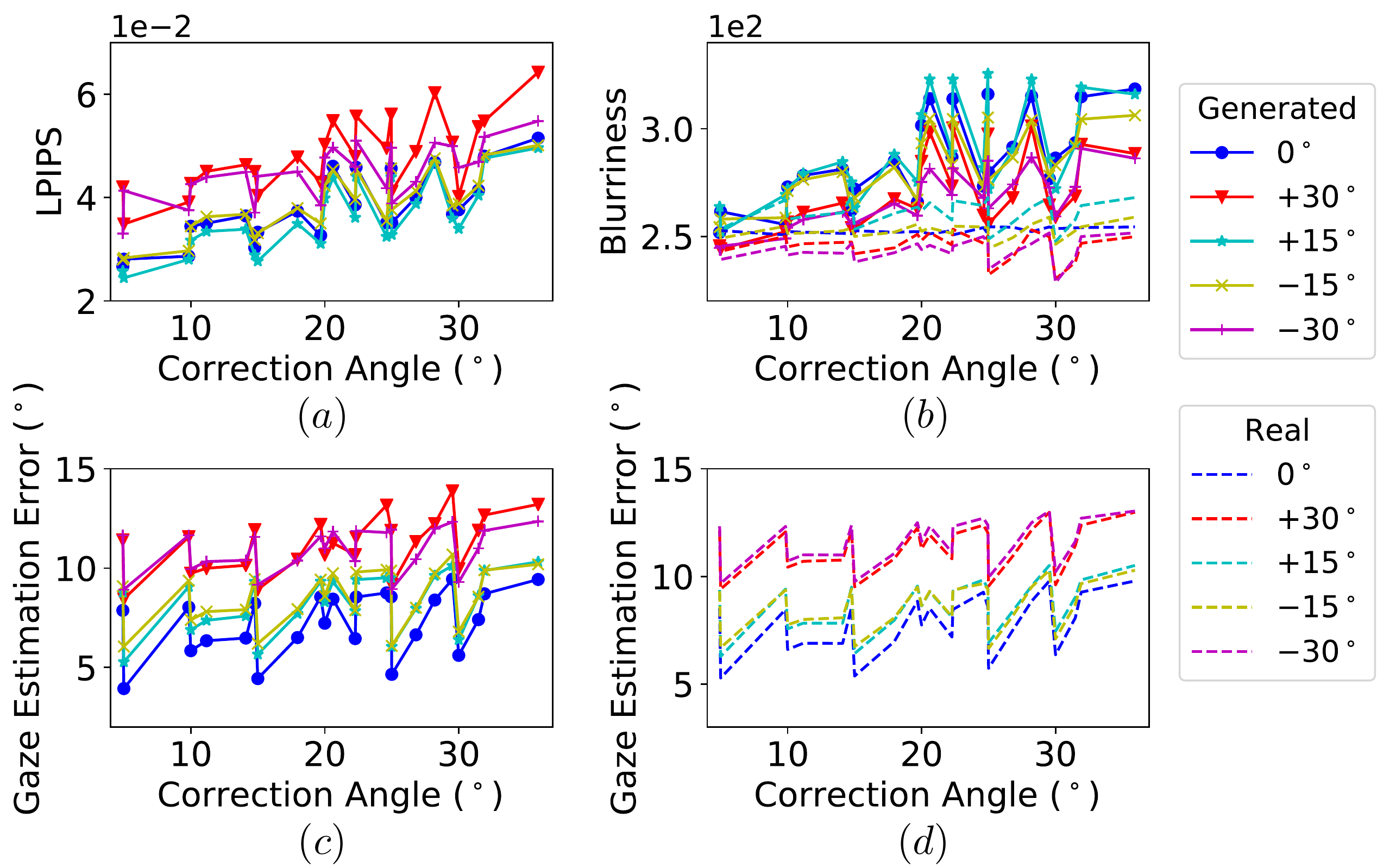}
\end{center}
   \caption{Quantitative evaluation results}
\label{fig:quantitative}
\end{figure}

Using the evaluation protocol and metrics introduced in \mbox{Sec. 5.1} and \mbox{Sec. 5.3} (in our paper), Fig. \ref{fig:quantitative}(a) shows that the LPIPS scores of the generated images are consistent up to $\pm15^\circ$.
The LPIPS scores for larger head angles ($\pm30^\circ$) are worse than the ones of ($0^\circ$, $\pm15^\circ$).
We note that: 1) There are fewer training samples with large head poses due to \texttt{dlib} detection failures. 2) These samples are more difficult in general, due to self-occlusion under extreme viewing angles. For example, in the input of the bottom row (Fig.~\ref{fig:qualitative}), the eye-corner is completely occluded by the nose.

The blurriness scores in Fig. \ref{fig:quantitative}(b) indicate that head pose only marginally affect image sharpness.

Fig. \ref{fig:quantitative}(c) shows that large head poses lead to large gaze estimation error for our generated images.
Comparing Fig. \ref{fig:quantitative}(c) and (d) shows that the gaze estimation error of generated and real images with the same head angle are consistent with each other.
It suggests that the generated images are of similar quality to real ones wrt to the gaze estimation task. 
In summary, this experiment provides evidence that the proposed method performs well, even on eye images generated with different head poses.

\end{document}